\documentclass[letterpaper, 10 pt, conference]{ieeeconf}  % Comment this line out if you need a4paper

\IEEEoverridecommandlockouts                              % This command is only needed if 
\usepackage{amsmath,amssymb,amsfonts}
\usepackage{algorithmic}
\usepackage{algorithm}
\usepackage{array}
\usepackage{textcomp}
\usepackage{stfloats}
\usepackage{url}
\usepackage{verbatim}
\usepackage{graphicx}
\usepackage{cite}
\usepackage{subfigure}
\usepackage{comment}
\usepackage{svg}
\usepackage{xcolor}

\title{\LARGE \bf
An Extended Horizon Tactical Decision-Making for Automated Driving Based on Monte Carlo Tree Search
}

\author{Karim Essalmi$^{1}$$^{,2}$, Fernando Garrido$^{1}$$^{,2}$, and Fawzi Nashashibi$^{1}$
\thanks{$^{1}$Inria Paris, 48 rue Barrault, 75013 Paris, France {\tt\small \{karim.essalmi, fernando.garrido-carpio, fawzi.nashashibi\}@inria.fr}}
\thanks{$^{2}$Valeo Mobility Tech Center, 6 rue Daniel Costantini, 94000 Créteil, France {\tt\small \{karim.essalmi, fernando.garrido\}@valeo.com}}}

\begin{document}

\maketitle
\thispagestyle{empty}
\pagestyle{empty}

%%%%%%%%%%%%%%%%%%%%%%%%%%%%%%%%%%%%%%%%%%%%%%%%%%%%%%%%%%%%%%%%%%%%%%%%%%%%%%%%

\begin{abstract}
This paper introduces COR-MCTS (Conservation of Resources - Monte Carlo Tree Search), a novel tactical decision-making approach for automated driving focusing on maneuver planning over extended horizons. Traditional decision-making algorithms are often constrained by fixed planning horizons, typically up to 6 seconds for classical approaches and 3 seconds for learning-based methods limiting their adaptability in particular dynamic driving scenarios. However, planning must be done well in advance in environments such as highways, roundabouts, and exits to ensure safe and efficient maneuvers. To address this challenge, we propose a hybrid method integrating Monte Carlo Tree Search (MCTS) with our prior utility-based framework, COR-MP (Conservation of Resources Model for Maneuver Planning). This combination enables long-term, real-time decision-making, significantly enhancing the ability to plan a sequence of maneuvers over extended horizons. Through simulations across diverse driving scenarios, we demonstrate that COR-MCTS effectively improves planning robustness and decision efficiency over extended horizons.
\end{abstract}

\section{Introduction}

\begin{figure}[!t]
    \centering    
    \subfigure[End of lane]{\includegraphics[width=.23\textwidth]{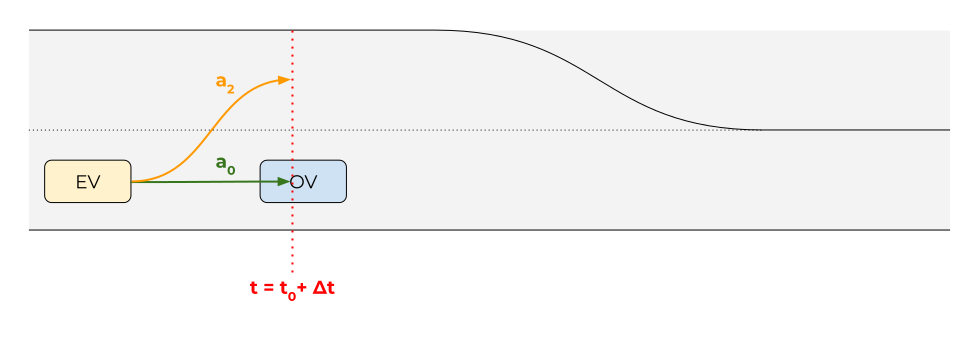}}
    \subfigure[Exit ramp]{\includegraphics[width=.23\textwidth]{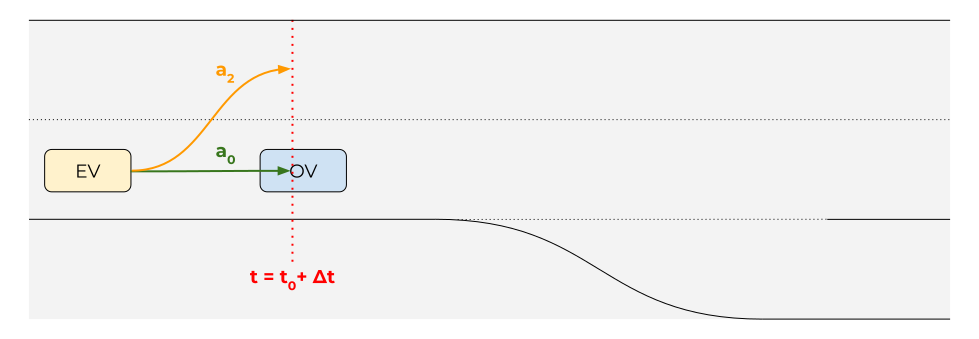}}\\
    \subfigure[Intersection]{\includegraphics[width=.46\textwidth]{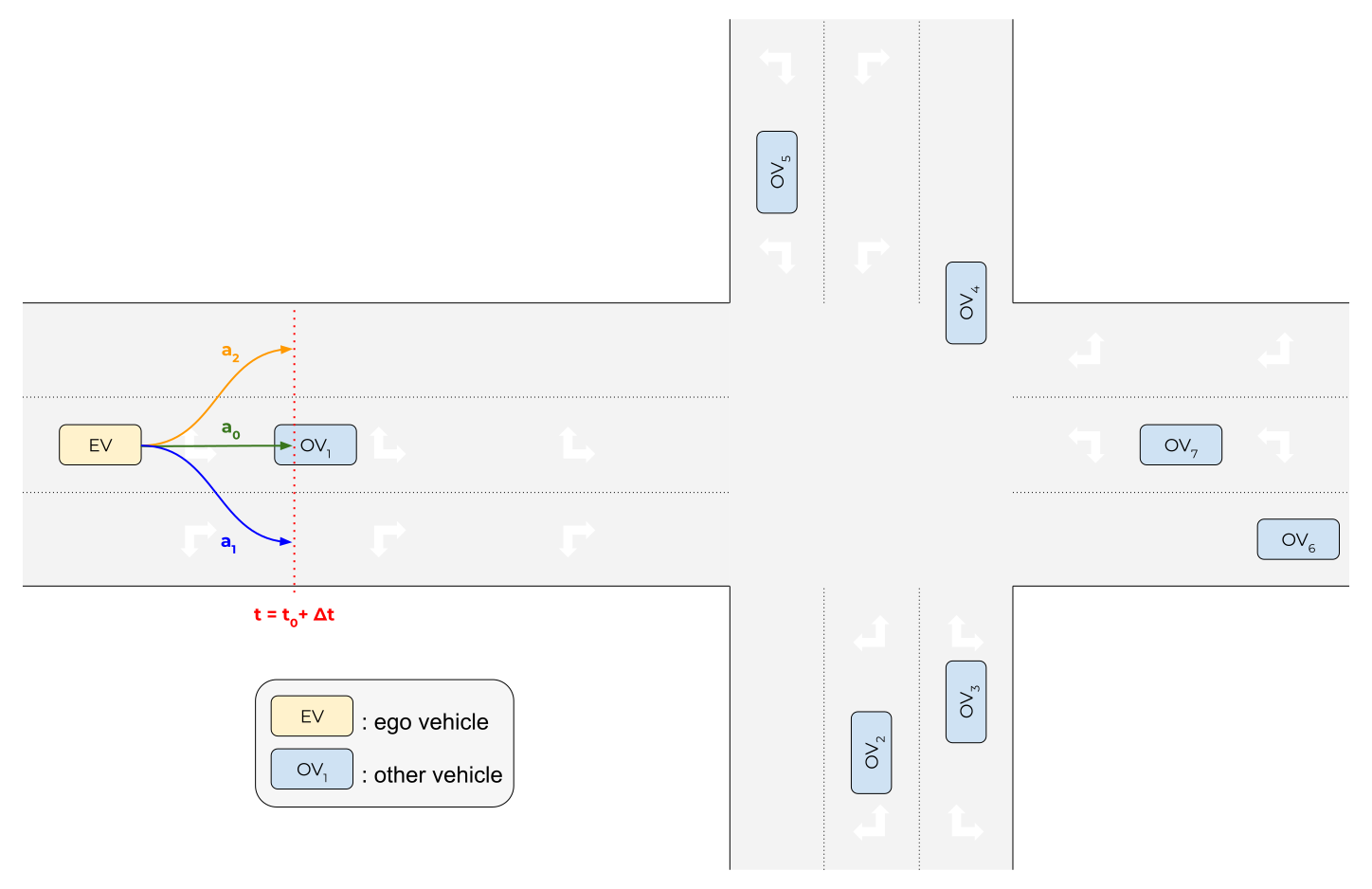}}\hfill   
    \caption{Examples of scenarios where a fixed planning horizon method might provide inappropriate outcomes. \(a_i\) represents the different possible actions: \(a_0\): Keep Lane action, \(a_1\): Change Lane Right action, \(a_2\): Change Lane Left action; \(t_0\) is the current time; \(\Delta t\) is the planning horizon. (a) End of lane: in this case, it is not advisable for the ego to move to the left lane, even though the vehicle ahead (OV) is driving slowly; otherwise, the EV could risk getting stuck in the left lane. (b): Exit-ramp: the ego should not overtake the vehicle, otherwise it will skip the exit lane. (c): Intersection: if the ego needs to turn to the left, it should not go back to the right lane otherwise it will be stuck and fail in its mission.}
    \label{fig:JustificationOfLongTermPlanning}
\end{figure}

% 1. Benefits of AV
% 2. (What is decision-making)???
% 3. Talk about the traditional method
% 4. Problematic (traditional method has proven their efficiency but there are still some issues, notably with planning in long term (like human does)). 
% 5. List briefly the traditional method to solve long-term decision-making (game tree, tree, MCTS, A*, ...). 
% 6. Talk about our technique (we combine MCTS with COR-MP): MCTS is known to solve decision-making problem implying long-term decision-making such as, in his first employment (AlphaGo and AlphaZero) to solve chess problem. Recently it have gain popularities to solve automated driving problems (cite papers using MCTS for DM in AD). But they mainly combined it using Machine Learning techniques.
% 7. Contributions
% 8. Structure of the paper 
The deployment of self-driving cars offers numerous benefits, such as improved transportation mobility, enhanced vehicle efficiency in terms of fuel consumption, and better traffic flow management \cite{othman2022exploring, bathla2022autonomous}. However, significant challenges remain before fully autonomous vehicles can be integrated into daily life. One of these challenges lies in the decision-making process. By leveraging information from the perception module (e.g., sensors, mapping), the decision-making component determines the sequence of actions the ego-vehicle (EV) must take to safely reach its destination. In the literature, decision-making is typically divided into different levels \cite{garrido2022review}: route planning, behavior planning, and trajectory planning. This paper focuses on behavior planning, also known as the tactical level. This layer ensures real-time decisions are safe, efficient, and comfortable, thus enhancing the user experience. Additionally, interpretability is becoming increasingly important, as it is crucial to understand the reasoning behind each decision.

To facilitate the adoption of autonomous vehicles alongside human-driven cars, it is essential to understand how humans make decisions while driving and to replicate their behavior in computational models. In our previous study, COR-MP \cite{CORMPConference}, we introduced a tactical decision-making framework inspired by the Conservation of Resources theory \cite{hobfoll1989}. This approach integrates human psychological principles into a computational model, enabling the EV to make decisions similar to those of a human driver.

Although recent works have tackled key limitations in existing decision-making algorithms; However, one challenge is still relevant: fixed planning horizon. This limitation can result in inappropriate behavior in certain scenarios, as illustrated in Figure \ref{fig:JustificationOfLongTermPlanning}. In fact, human drivers, in contrast, often make decisions well in advance while driving, justifying the importance of considering longer planning horizons. For instance, at the tactical level of decision-making, conventional algorithms typically operate with a planning horizon of up to six seconds \cite{garrido2022review}. While this can be sufficient in many cases, it fails to adapt effectively in more complex or dynamic situations. 

To address the challenge of long-term decision-making, the problem is often modeled following a tree structure, allowing for the evaluation of multiple future scenarios. Various approaches are used to solve this tree-based problem: game-theoretic methods, as proposed in \cite{saraoglu2023minimax, fisac2019hierarchical}; brute force approaches, which explicitly model every possible action, as seen in \cite{li2017explicit}, though these can become computationally expensive as the state-space and planning horizon expand; or Monte Carlo Tree Search methods, as proposed in \cite{chekroun2024mbappe, wen2024monte, karimi2020receding}.

MCTS is a heuristic tree search algorithm designed to solve decision-making problems. It combines the Monte Carlo method - a technique that involves repeated random sampling - with a tree search framework. Originally introduced in \cite{coulom2006efficient}, MCTS was proposed as a solution for developing computer players for the game of Go. More recently, it has garnered significant attention, particularly through two groundbreaking studies by Google DeepMind: AlphaGo \cite{silver2016mastering} and AlphaZero \cite{silver2017mastering}. These studies leverages the strengths of both deep learning and MCTS techniques. The first, AlphaGo, used supervised learning to train a computer player for the game of Go, which is known for its vast number of possible moves. The second, AlphaZero, extended the approach to a variety of games, including chess, shogi, and Go, by combining MCTS with reinforcement learning, without relying on a dataset, thus using unsupervised learning.

The main contribution of this work is the extension of COR-MP to COR-MCTS to make it more human-like by integrating the Monte Carlo Tree Search technique. This combination enhances the planning horizon, allowing for the evaluation of a sequence of actions. This enables long-term planning and aligns the approach more closely with human decision-making while driving.
% Contributions
%The main contributions % a solution to solve some limitations of traditional methods by employing MCTS. A framework allowing long-term human-like decision-making for maneuver planning application in automated driving domain. 
% Combination of our previous method whish is a utility-based technique with MCTS allowing long-term decision-making

% Structure of the paper
The remainder of the paper is structured as follows. Section \ref{section::relatedWork} reviews related work on long-term decision-making in the automated driving domain. Our approach is described in Section \ref{section::method}, followed by a presentation of the obtained results in Section \ref{section::results}. Finally, conclusions and future research directions are detailed in Section \ref{section::conclusion}.

% Section - Related Work
\section{Related work}\label{section::relatedWork}

Some studies have already considered evaluating sequences of actions, rather than just individual actions, to extend the planning horizon \cite{gonzalez2019human, zhang2020efficient}. Typically, the problem is modeled as a tree, where the main objective is to solve this tree-structured decision-making challenge. 

% POMDP
For instance, the authors in \cite{zhang2020efficient} modeled the problem as a Partially Observable Markov Decision Process (POMDP), allowing to account for the uncertainty introduced by the world representation. They solved this model using a DCP-Tree (domain-specific closed-loop policy) technique, which helps guide the tree search. Their approach plans for 8 seconds, with an action time duration of 2 seconds. Similarly, \cite{li2017explicit} also modeled the problem as a POMDP but solved it using a standard decision tree. While modeling the environment through POMDP can define probabilistic transitions and observations, it requires accurate models, which often results in high computational costs \cite{nelson}. This can contradict the need for real-time decision-making, which must be reactive. To address this, \cite{gonzalez2019human} combined POMDP with the Monte Carlo Tree Search technique to reduce computational complexity, as MCTS can reduce it. In their approach, they consider two actions: lane change and lane keeping, and managed acceleration using the Intelligent Driver Model (IDM) \cite{IDM}.

% Game theory
Another method for modeling long-term decision-making is through game theory, as demonstrated by the authors in \cite{saraoglu2023minimax}. They focused on highway driving and modeled the problem as a two-player game, with the environment represented as a game tree, which is solved using the minimax technique. A notable advantage of modeling the problem as a game is that it makes the model interaction-aware. Consequently, it also increases computational complexity as the number of players or strategies grows.
In \cite{fisac2019hierarchical}, the authors considered a hierarchical game model where short-term decisions are influenced by long-term predictions, specifically for motion planning applications. 

% IDM and MOBIL
The authors in \cite{moghadam2021autonomous} proposed a long-term maneuver planning framework that combines IDM \cite{IDM} and Minimizing Overall Braking Induced by Lane changes (MOBIL) \cite{MOBIL} to handle both longitudinal and lateral ego movements. By tuning these two models, the authors successfully model three different driver profiles: agile, moderate, and conservative, and simulate their behavior in long-term planning. 

% MCTS
% 1. First talk about alphaGO and alphaZero // Or move this to INTRO ???
% 2. Then say that it can be applied in automated driving
% 3. Paper from raphael say that it is for motion planning application
% 4. The one from arxiv (utility and MCTS): similar to us but the planning horizon is lower.

%Talk about alphaGo and AlphaZero (which are the first using MCTS).
%Say that MCTS is mainly using by combining it with ML approach but it can be used with traditional method (utility, rule-based, or others). 

%MCTS was firslty used for game and then, we found at that it can be applied for decision-make, and particularly in automated driving domain. (ref to papers using MCTS for Decision-making in automated driving)

Finally, in the literature, the most common approach for solving long-term decision-making problems is through the Monte Carlo Tree Search technique. For example, in \cite{karimi2020receding}, the authors solve a motion planning problem by leveraging MCTS and level-k game techniques for lane change and merge scenarios in highway situations. As demonstrated in the groundbreaking works of \cite{silver2016mastering}, and \cite{silver2017mastering}, combining MCTS with learning techniques is a powerful approach. In \cite{chekroun2024mbappe}, MCTS is combined with deep learning to solve motion planning problems in real time. The study proposed in \cite{wen2024monte} combines MCTS with a utility-based technique for behavior planning applications, handling various scenarios such as lane changes, acceleration, and deceleration situations.

% Resume of the literature review - Why we choose MCTS
As demonstrated in the literature, the Monte Carlo Tree Search method efficiently addresses long-term decision-making challenges. By evaluating sequences of actions through a tree structure, MCTS can extend the planning horizon. Furthermore, computational complexity is a crucial factor for real-time decision-making, and MCTS offers acceptable computational performance since the tree search can be terminated at any time. Lastly, planning over a longer horizon inherently introduces more uncertainties, which MCTS can effectively manage due to its stochastic nature.

\begin{figure}[!t]
\centering
\includegraphics[width=0.5\textwidth]{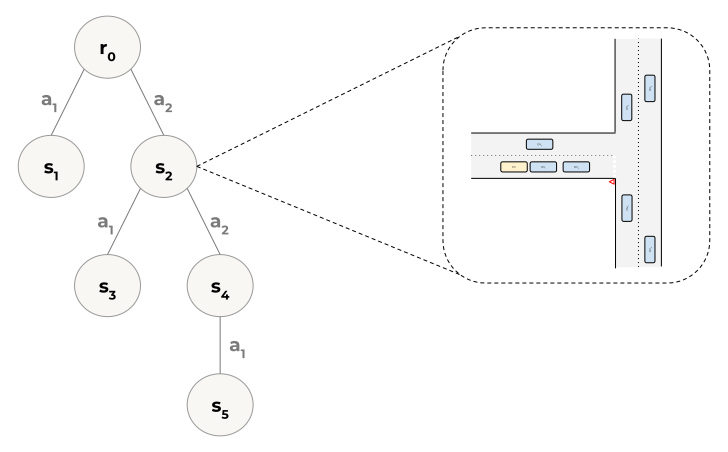}
\caption{Example of a tree structure illustrating how a state is defined. \(r_0\): root node, representing the current state of the system; \(a_i\): action \(i\); \(s_i\): state \(i\).}
\label{fig:nodes_mcts}
\end{figure}

% Section - Method
\section{Method}\label{section::method}
% Figure method
\begin{figure*}[ht]
\includegraphics[width=\textwidth]{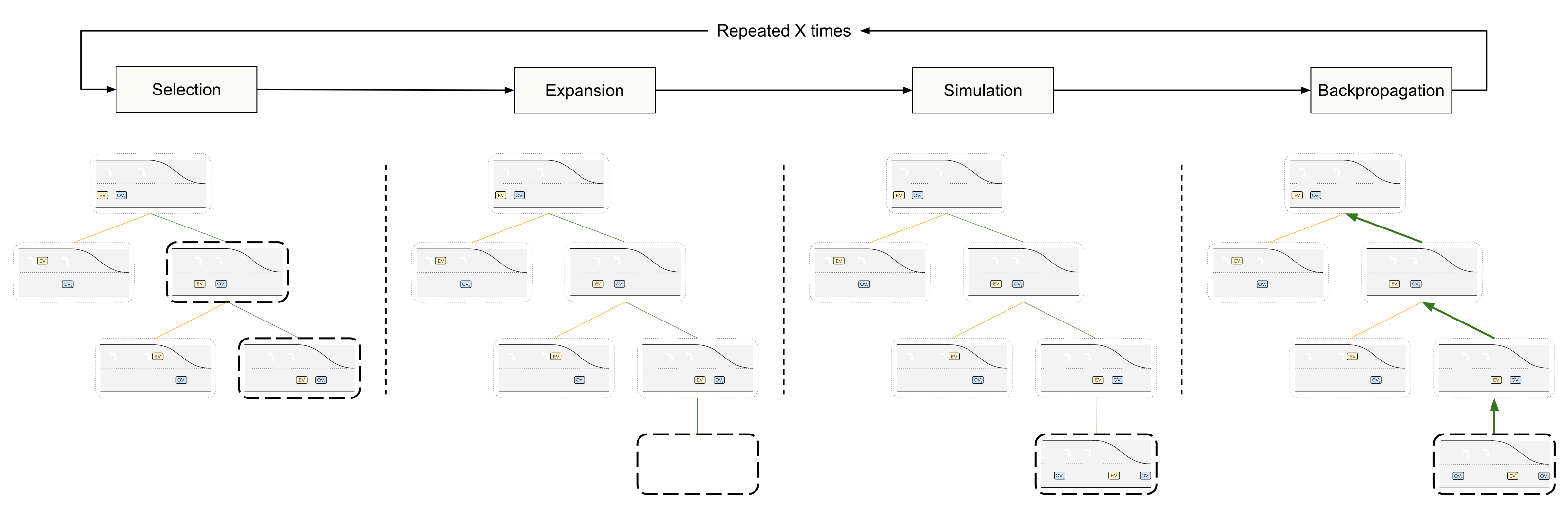}
\caption{Steps of our method, where orange edges represent a change lane left action and green ones represent a keeping lane action. For each predicted state \(s_k\), our method assigns a \(v\) value reflecting the benefit of being in \(s_k\). \(v\) is then backpropagates until the root node, hence short-term decisions are influenced by long-term ones. These steps are repeated X times until a suitable solution is found.}
\label{fig:mcts_steps}
\end{figure*}

MCTS is typically modeled using nodes \(\mathcal{N}\), which represent the diverse states \(\mathcal{S}\) of the system, and edges \(\mathcal{V}\), which link states and represent the action \(a \in \mathcal{A}\) leading to new states \(s_{k+1} \in \mathcal{S}\) from previous states \(s_{k} \in \mathcal{S}\). Figure \ref{fig:nodes_mcts} illustrates an example of a tree structure used to model the problem. 

In our method, each node \(n \in \mathcal{N}\) is defined as a tuple \( \langle v, m, U, UCB \rangle \), where: 
\begin{itemize}
    \item \(v\) represents the immediate profit value of a state, a metric that quantifies how favorable it is to be in the evaluated state,
    \item \(m\) is the number of visits to node \(n\),
    \item \(U\) is the accumulated profit value, and
    \item \(UCB\) is the Upper Confidence Bound value.
\end{itemize}

As illustrated in Figure \ref{fig:mcts_steps}, the MCTS method consists of four distinct steps. In the following, we provide a detailed explanation of our approach, which integrates the Monte Carlo Tree Search technique.

% Subsection - Selection
\subsection{Selection}
Starting from the root node \(r\), child selection is recursively applied until a leaf node \(l\) (a node with no children) is reached. This leaf node will later be expanded and simulated based on the information gathered by the system. The selection of \(l\) is crucial, as it guides the search within the tree. A balance must be made between exploration (investigating unexplored branches that may lead to better or worse states) and exploitation (leveraging the current knowledge of the world and exploiting it as is). 

Various selection strategies have been proposed in the literature, including PUCT (Predictor + Upper Confidence bounds applied to Trees) \cite{rosin2011multi}, EXP3 (Exponential-weight algorithm for Exploration and Exploitation) \cite{exp3}, and the widely used UCT (Upper Confidence Bounds applied to Trees) \cite{kocsis2006bandit}, which extends the UCB algorithm from \cite{auer2002finite}. In our method, node selection is performed using the UCB technique, as defined in Equation \ref{eq:ucb}.

\begin{equation}
\label{eq:ucb}
    UCB(n)=\Tilde{U}(n)+c \cdot \sqrt{\frac{ln(M)}{m(n)}}
\end{equation}

Where \(\Tilde{U}(n)\), \(c\), \(M\), and \(m(n)\) represent, respectively: the mean of the accumulated profit value, a constant balancing exploration and exploitation (set to \(c=\sqrt{2}\)), the number of visits to the parent node, and the number of visits to the corresponding node.

% Subsection - Expansion
\subsection{Expansion}
At this step, at least one child node is added to the tree, unless \(l\) is a terminal state. The new node represents the state \(s_{k+1}\), which is assumed to be reached by performing an action \(a\). We define the ego action space as \(a \in \mathcal{A}\), where: 

\begin{equation}
    \mathcal{A} =     
    \left\{
    \begin{array}{c}
        \text{Change\ Lane\ Left} \\
        \text{Change\ Lane\ Right} \\
        \text{Keep\ Lane\ Accelerate} \\
        \text{Keep\ Lane\ Same\ Speed} \\
        \text{Keep\ Lane\ Decelerate} \\
        \text{Stop}
    \end{array}
    \right\}    
\end{equation}

Thus, we account for both longitudinal and lateral ego movements. The action to be simulated is chosen based on the action leading to the parent node. Specifically, if the parent node's action involved a lane change, expansion slightly favors keeping lane maneuvers. This strategy avoids consecutive lane change maneuvers. In all other cases, the expansion action is selected randomly following a uniform distribution over \(\mathcal{A}\).

% Subsection - Simulation
\subsection{Simulation}
The goal of this step is twofold. First, to simulate the environment, including the ego-vehicle and interacting agents, based on the chosen action \(a\). Second, to evaluate the new simulated state by quantifying how favorable it is to be in this state. This is achieved by applying our previous work, COR-MP \cite{CORMPConference}.
During the simulation, the system predicts what will happen if the EV performs the evaluated maneuver. A profit value \(v \in [0,1]\) is then assigned to the new state \(s_{k+1}\). This value serves as a metric that quantifies the benefit of being in \(s_{k+1}\), taking into account various factors such as driving comfort, safety, adherence to traffic rules, and more. A detailed explanation of how this metric is computed can be found in our previous study \cite{CORMPConference}. Finally, the objective is to maximize \(v\): a low value indicates that \(s_{k+1}\) is undesirable and should be avoided, whereas a high value suggests the state is favorable. 

% Subsection - Backpropagation
\subsection{Backpropagation}
At last, the payoff is backpropagated to the root node (Equation \ref{eq:backpropagation}), following the path that led to \(s_{k+1}\). In this way, immediate actions (short-term) are influenced by deeper nodes in the tree, enabling long-term decision-making. 

\begin{equation}
\label{eq:backpropagation}
    U(n) \gets U(n) + \gamma^t \cdot v(n)
\end{equation}

Here, \(U(n)\), \(\gamma\), and \(t\) represent, respectively, the accumulated profit value of \(n\), the discount factor (set to \(\gamma = 0.9\)), and the depth of a node relative to the root node \(r\) (t=0 for the leaf node \(l\)).

\subsection{Decision}
As illustrated in Figure \ref{fig:mcts_steps}, the previous steps are repeated \(X\) times until a satisfactory solution is found within an acceptable computational time. During our experiments, we stopped the tree search either when the computational time exceeded one second or when the number of nodes in the tree reached 50. 

Finally, the decision to be made instantly by the EV is the action that leads to the node \(n^*\) (Equation \ref{eq:af}).

\begin{equation}
\label{eq:af}
    n^* = \arg\max_{n \in \mathcal{N}_r} U(n)
\end{equation}

With \(\mathcal{N}_r\) denoting the set of nodes that are direct children of the root node \(r\) (i.e., the first row of the tree). 

% Section - Results
\section{Results}\label{section::results}

% Figure Results
\begin{figure*}
\centering
\subfigure[Scenario 1: End of left lane.]{\includegraphics[width=.48\textwidth]{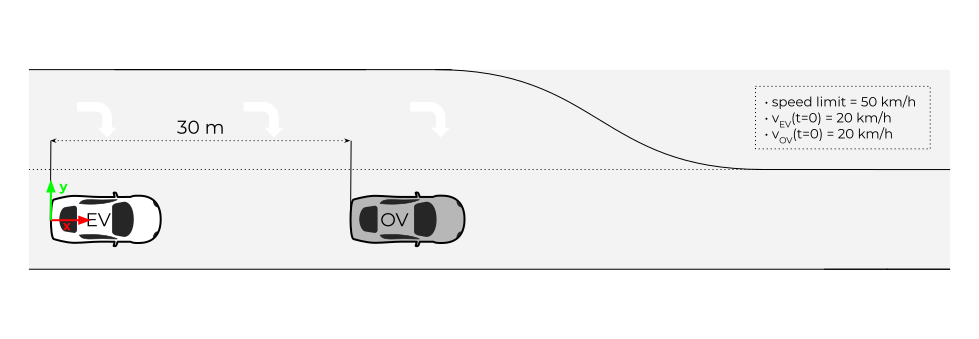}}
\vspace{0.2cm}
\subfigure[Scenario 2: Approaching an exit lane.]{\includegraphics[width=.48\textwidth]{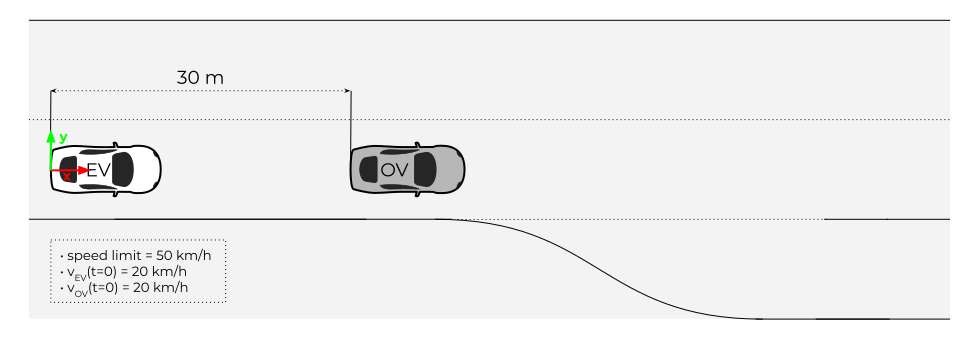}}\\
\vspace{0.2cm}
\subfigure[Different outcomes observed while running Scenario 1.]{\includegraphics[width=\textwidth]{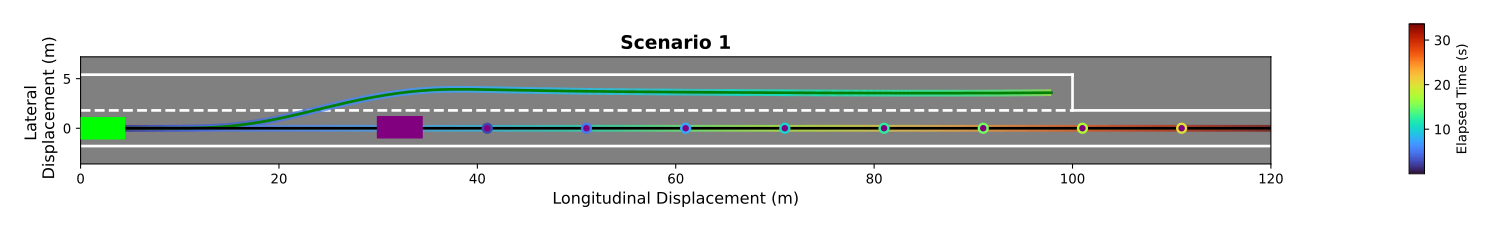}}\\
\subfigure[Different outcomes observed while running Scenario 2.]{\includegraphics[width=\textwidth]{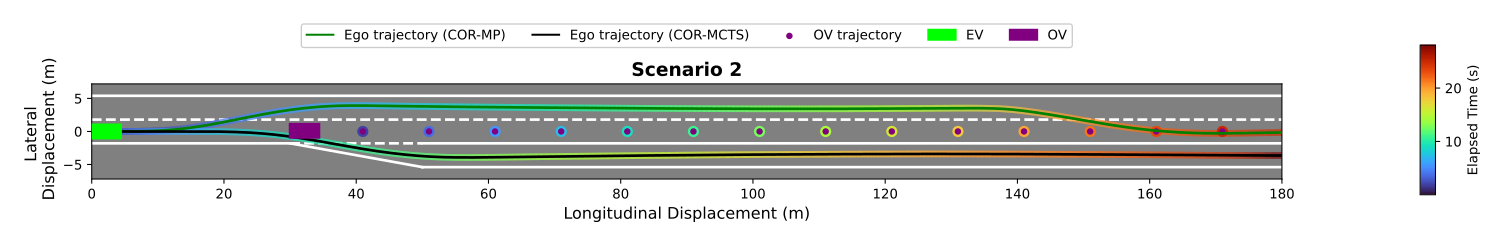}}\\
\vspace{0.2cm}
\subfigure[Distribution of runtime for different approaches, visualized using violin plots, obtained from multiple simulations.]{\includegraphics[width=.9\textwidth]{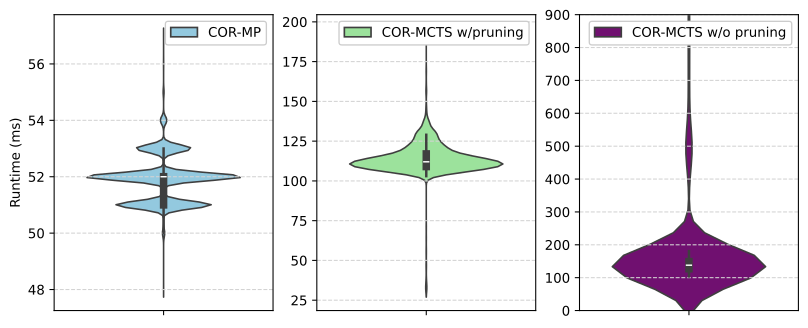}}\\
%{figures/Results/time_complexity_hist.png}}\\
%{figures/Results/time_complexity_hist_2.png}}\\
\caption{Various results obtained by testing our approach. Subfigures (a) and (b) illustrate the two scenarios presented in this paper. Subfigures (c) and (d) highlight the behavior adopted by the ego-vehicle by running the two presented scenarios for two different algorithms. Subfigure (e) shows the repartition of the runtimes of our previous approach (COR-MP), and the ones of the method presented in this study: COR-MCTS with (w/) pruning and COR-MCTS without (w/o) pruning. These runtime values have been obtained through various simulated scenarios.}
\label{fig:Results}
\end{figure*}

% Figure - Screenshots
\begin{figure*}
\vspace{0.2cm}
\centering
\subfigure[Scenario 1 - COR-MP]{\includegraphics[width=0.24\textwidth]{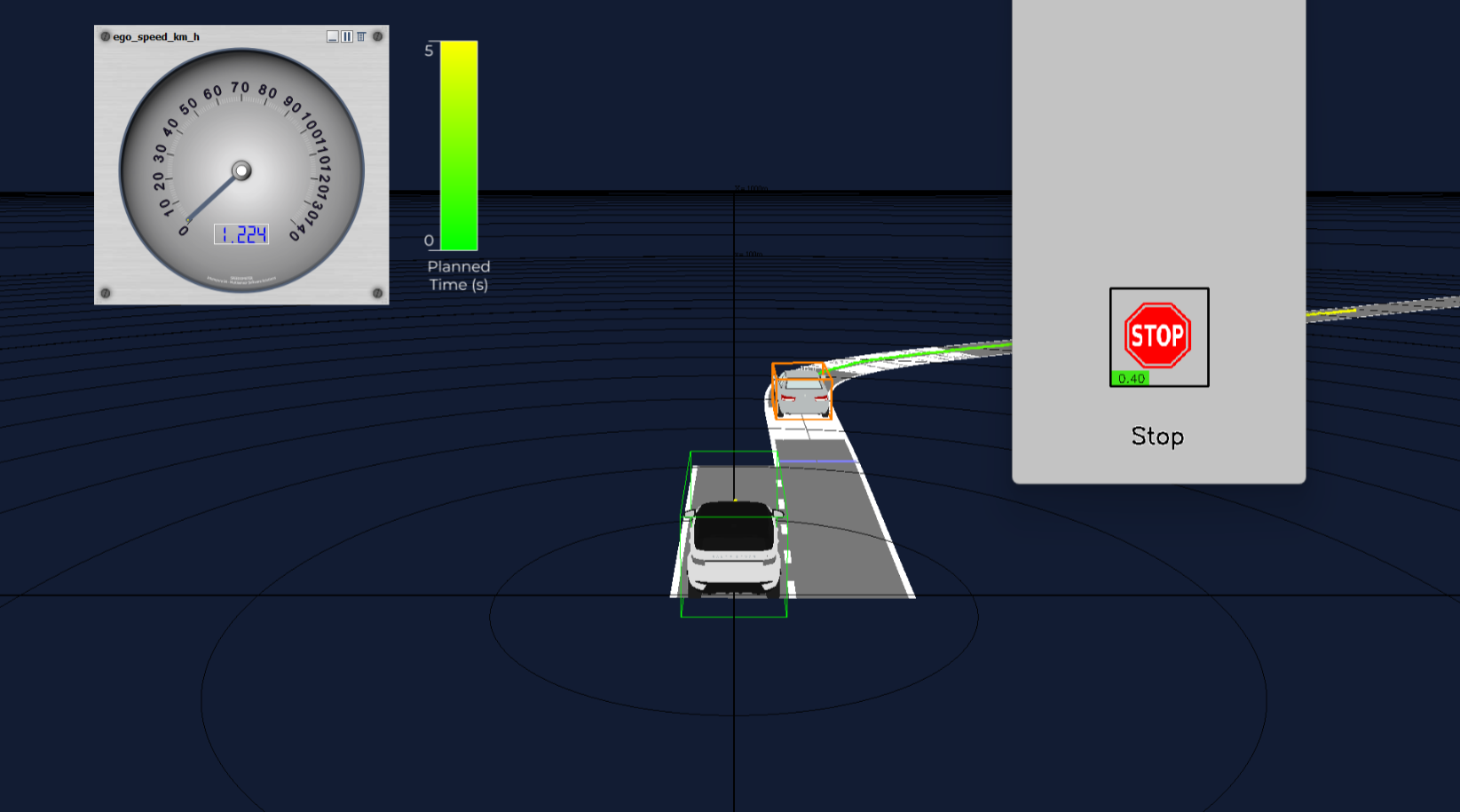}}
\subfigure[Scenario 1 - COR-MCTS]{\includegraphics[width=0.24\textwidth]{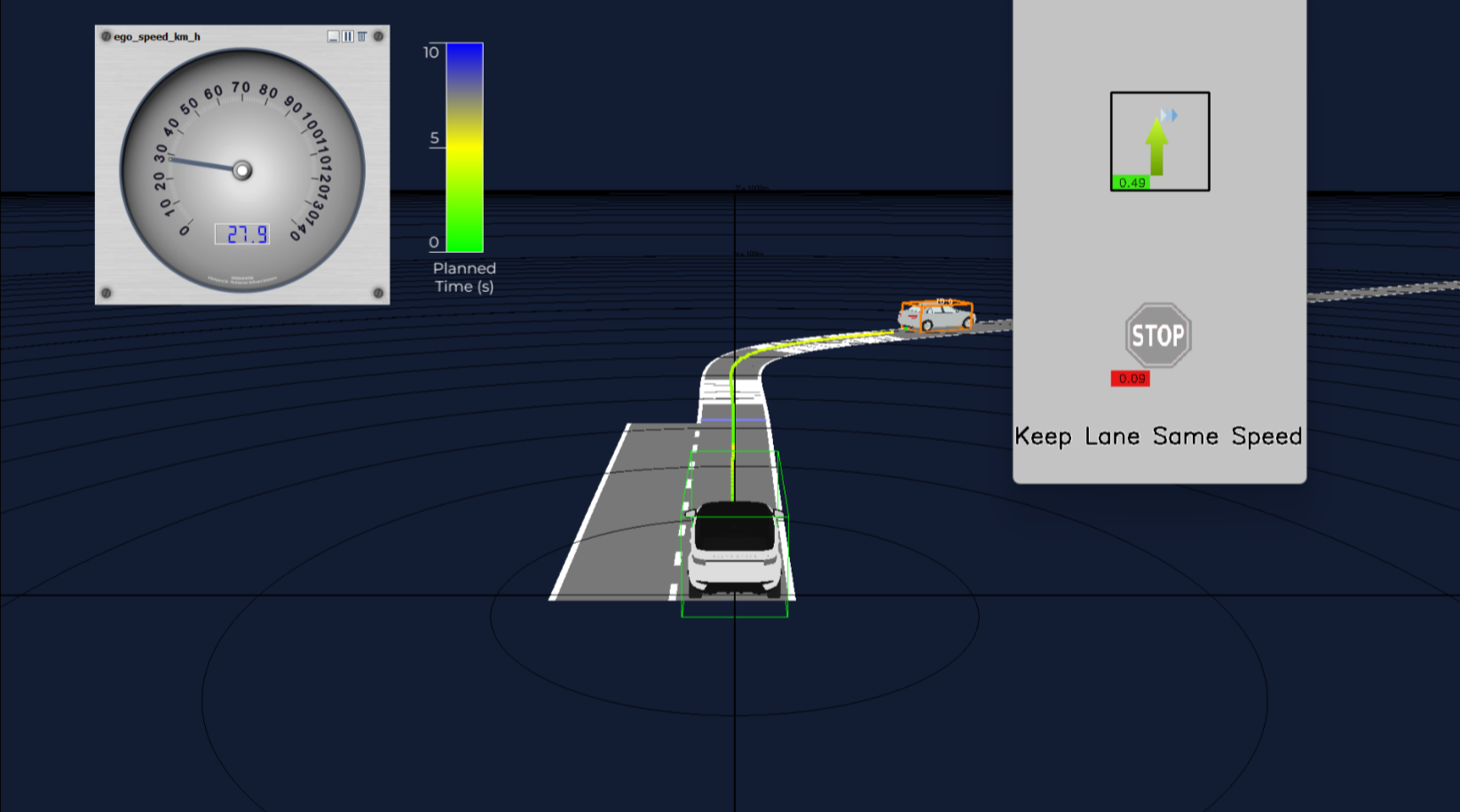}}
\subfigure[Scenario 2 - COR-MP]{\includegraphics[width=0.24\textwidth]{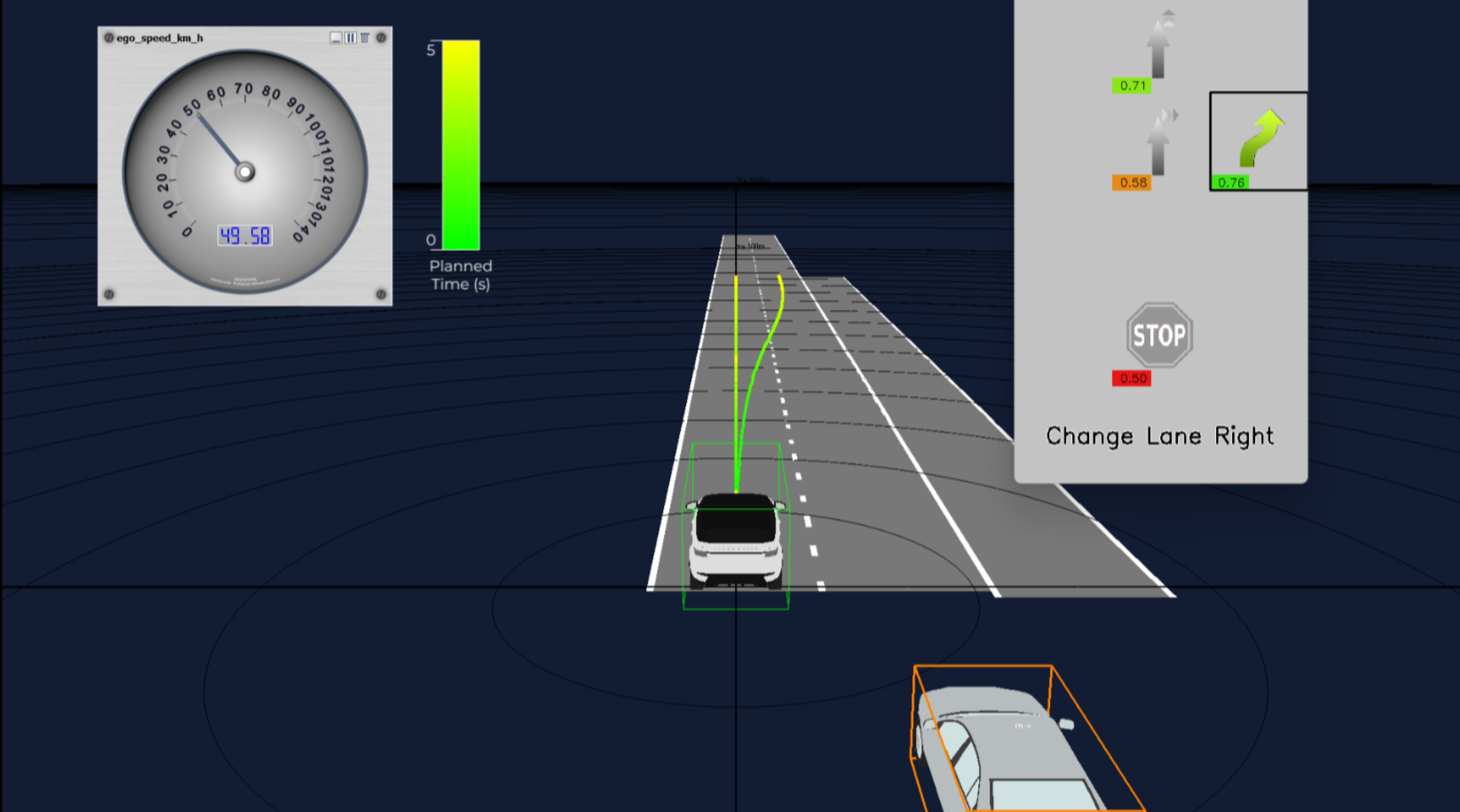}}
\subfigure[Scenario 2 - COR-MCTS]{\includegraphics[width=0.24\textwidth]{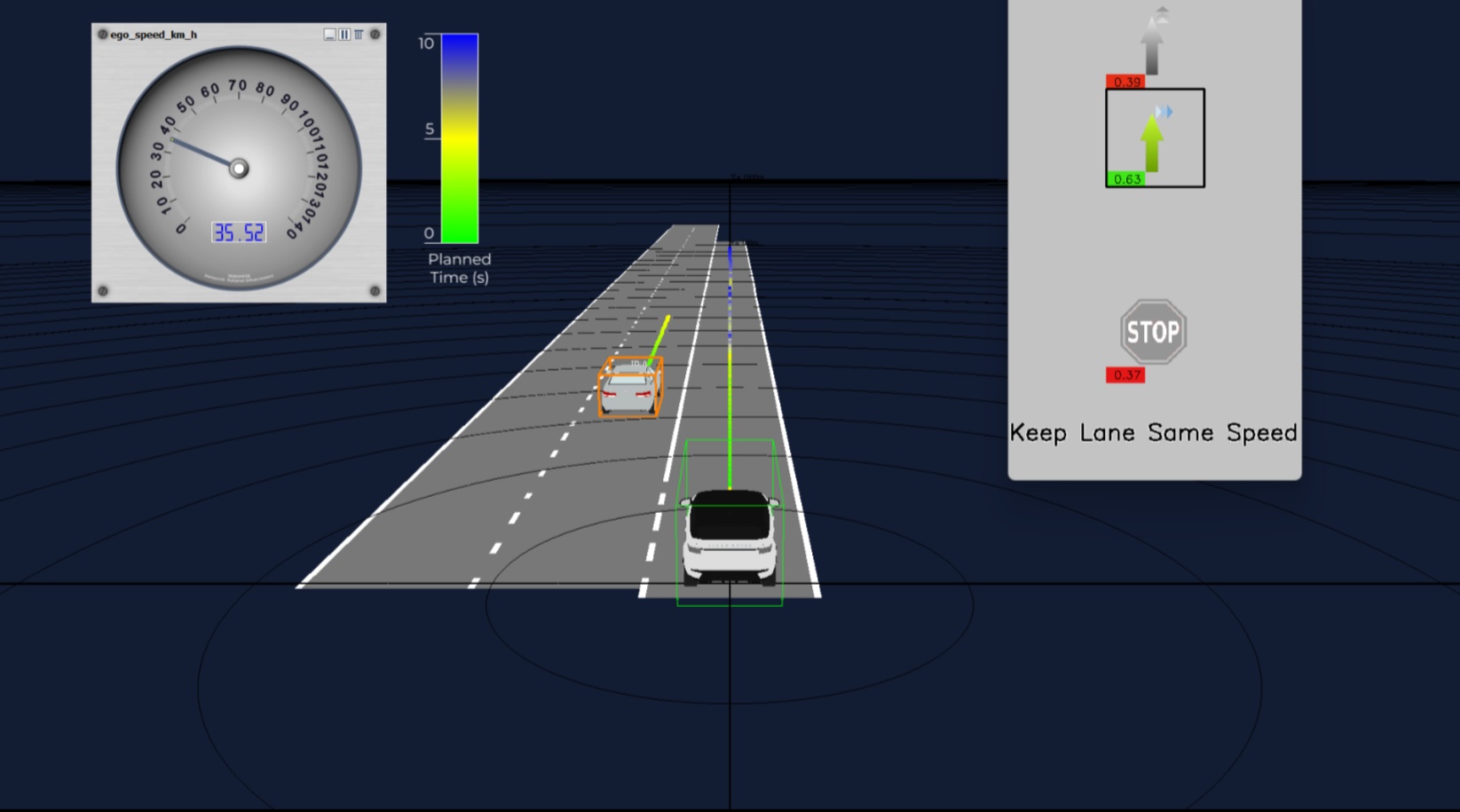}}\\
\caption{Diverse screenshots captured while running the two scenarios presented in this study using our simulator. In each screenshot, the current ego vehicle speed is displayed on the gauge in the top-left corner, while the color label represents the planning time for the displayed ego trajectories. The top-right viewer in each subfigure shows the feasible maneuvers, their assigned values \(v\), and the maneuver selected by the maneuver planner, which is highlighted with a black square. Subfigures (a) and (c) depict moments where the classical method produces inappropriate outcomes for Scenario 1 and Scenario 2, respectively. In contrast, subfigures (b) and (d) illustrate these situations using an extended planning horizon method, demonstrating its improved decision-making.}
\label{fig:Screenshots}
\end{figure*}

% Presenting the scenarios
Our approach has been tested through simulation using our own simulator, which is based on the RTMaps middleware \cite{RTMapsFawzi}. Figure \ref{fig:Screenshots} shows the simulator viewer. First, we tested our method in basic scenarios to ensure that the proposed decisions are safe, collision-free, and adhere to traffic rules. Then, we compared COR-MCTS to classical maneuver planning methods. Here, we present two scenarios in which traditional methods, based on a fixed planning horizon, often yield inappropriate outcomes.
The first scenario, illustrated in subfigure (a) of Figure \ref{fig:Results}, involves a two-lane road where the left lane ends. An interacting vehicle ahead of the ego vehicle is driving below the speed limit. The second scenario is also a two-lane road situation, where the ego vehicle is approaching an exit lane. In this scenario, another interacting vehicle, also driving below the speed limit, is present ahead of the ego vehicle. In both scenarios, the speed limit is set to 50 km/h, the interacting vehicle's speed is set to 20 km/h, and the ego vehicle starts at 20 km/h.

% Talking about the results
Subfigures (c) and (d) of Figure \ref{fig:Results} show the decisions obtained, particularly the path followed by the EV during the test. The green trajectory represents the path obtained using a traditional approach \cite{CORMPConference} with a fixed planning horizon of 5 seconds, while the black trajectory corresponds to the path followed by employing the approach presented in this study. The purple points indicate the positions of the interacting vehicle over time. Finally, around each path, there is a color fade indicating the elapsed time during the simulation.

As highlighted in these figures, the two algorithms behave completely differently. In the first scenario, the fixed planning horizon method decides to overtake the slower vehicle to reach the speed limit. However, this results in the ego vehicle getting stuck in the left lane, as it does not have enough time to safely merge back to the right lane. Subfigure (a) of Figure \ref{fig:Screenshots} illustrates this moment. In contrast, by employing an extended planning horizon through MCTS, the algorithm anticipates the end of the left lane and chooses to maintain its current speed, even though it remains below the speed limit. As shown by the subfigure (b) of Figure \ref{fig:Screenshots}.

In the second scenario, the traditional method behaves similarly: it prefers to overtake the slower vehicle to reach the speed limit. As a result, the ego vehicle is no longer able to take the exit lane. As shown in subfigure (c) of Figure \ref{fig:Screenshots}. In contrast, with MCTS, the algorithm is aware that the ego vehicle is approaching an exit lane. Therefore, as in the first scenario, the extended planning horizon approach opts to maintain the current speed, ensuring that the ego vehicle can take the exit. This is illustrated in subfigure (d) of Figure \ref{fig:Screenshots}. Finally, subfigure (e) of Figure \ref{fig:Results} presents the distribution of different runtimes obtained from different simulations to validate the proposed method. The blue plot represents the distribution of runtimes from our previous approach, COR-MP, while the green and purple plots correspond to COR-MCTS with pruning and COR-MCTS without pruning, respectively. We observed the following median runtime values: 51.9 ms for COR-MP, 113.88 ms for COR-MCTS with pruning, and 195.56 ms for COR-MCTS without pruning, which are suitable for maneuver planning applications, as they need to be less reactive than motion planning. However, we observed that at certain moments, while running COR-MCTS without pruning, the algorithm can get stuck in the tree, causing the runtime to increase significantly. By incorporating pruning functions, these situations can be mitigated, making MCTS more efficient for reactive and real-time applications, such as those required for maneuver planning.

% Section - Conclusion
\section{Conclusion}\label{section::conclusion}
This study presents a novel approach that combines Monte Carlo Tree Search with a utility-based maneuver planner: COR-MCTS. This integration enables long-term planning and addresses some limitations of existing approaches that rely on a fixed planning horizon. By extending the planning horizon, the Monte Carlo Tree Search technique helps prevent the “frozen robot” phenomenon, making the decision-making algorithm more adaptable to diverse scenarios. However, increasing the planning horizon results in higher computational complexity and introduces greater uncertainty while predicting future states. Although pruning is an effective technique for reducing computational time, it can lead to suboptimal solutions. Ultimately, a trade-off between computational efficiency and solution optimality must be carefully considered to ensure a reliable and effective decision-making model. In future work, we first plan to test this approach on a real vehicle to validate it under real-world conditions. Then, we aim to incorporate interactions with other road users in the decision-making process, making the approach more interaction-aware.

%\section{Acknowledgments}
% Bibliography
\bibliographystyle{IEEEtran}
%\bibliography{bibliography}

\begin{thebibliography}{10}
\providecommand{\url}[1]{#1}
\csname url@samestyle\endcsname
\providecommand{\newblock}{\relax}
\providecommand{\bibinfo}[2]{#2}
\providecommand{\BIBentrySTDinterwordspacing}{\spaceskip=0pt\relax}
\providecommand{\BIBentryALTinterwordstretchfactor}{4}
\providecommand{\BIBentryALTinterwordspacing}{\spaceskip=\fontdimen2\font plus
\BIBentryALTinterwordstretchfactor\fontdimen3\font minus \fontdimen4\font\relax}
\providecommand{\BIBforeignlanguage}[2]{{%
\expandafter\ifx\csname l@#1\endcsname\relax
\typeout{** WARNING: IEEEtran.bst: No hyphenation pattern has been}%
\typeout{** loaded for the language `#1'. Using the pattern for}%
\typeout{** the default language instead.}%
\else
\language=\csname l@#1\endcsname
\fi
#2}}
\providecommand{\BIBdecl}{\relax}
\BIBdecl

\bibitem{othman2022exploring}
K.~Othman, ``Exploring the implications of autonomous vehicles: A comprehensive review,'' \emph{Innovative Infrastructure Solutions}, 2022.

\bibitem{bathla2022autonomous}
G.~Bathla, K.~Bhadane, R.~K. Singh, R.~Kumar, R.~Aluvalu, R.~Krishnamurthi, A.~Kumar, R.~Thakur, and S.~Basheer, ``{Autonomous vehicles and intelligent automation: Applications, challenges, and opportunities},'' \emph{Mobile Information Systems}, vol. 2022, no.~1, p. 7632892, 2022.

\bibitem{garrido2022review}
F.~Garrido and P.~Resende, ``Review of decision-making and planning approaches in automated driving,'' \emph{IEEE Access}, vol.~10, 2022.

\bibitem{CORMPConference}
K.~Essalmi, F.~Garrido, and F.~Nashashibi, ``{COR-MP: Conservation of Resources Model for Maneuver Planning},'' in \emph{2024 IEEE 20th International Conference on Intelligent Computer Communication and Processing (ICCP)}.\hskip 1em plus 0.5em minus 0.4em\relax IEEE, 2024, pp. 1--8.

\bibitem{hobfoll1989}
S.~E. Hobfoll, ``{Conservation of resources: A new attempt at conceptualizing stress.}'' \emph{American psychologist}, vol.~44, no.~3, p. 513, 1989.

\bibitem{saraoglu2023minimax}
M.~Saraoglu, H.~Jiang, M.~Schirmer, {\.I}.~Mutlu, and K.~Janschek, ``{A Minimax-Based Decision-Making Approach for Safe Maneuver Planning in Automated Driving},'' in \emph{2023 American Control Conference (ACC)}.\hskip 1em plus 0.5em minus 0.4em\relax IEEE, 2023, pp. 4683--4690.

\bibitem{fisac2019hierarchical}
J.~F. Fisac, E.~Bronstein, E.~Stefansson, D.~Sadigh, S.~S. Sastry, and A.~D. Dragan, ``Hierarchical game-theoretic planning for autonomous vehicles,'' in \emph{2019 International conference on robotics and automation (ICRA)}.\hskip 1em plus 0.5em minus 0.4em\relax IEEE, 2019, pp. 9590--9596.

\bibitem{li2017explicit}
N.~Li, H.~Chen, I.~Kolmanovsky, and A.~Girard, ``An explicit decision tree approach for automated driving,'' in \emph{Dynamic systems and control conference}.\hskip 1em plus 0.5em minus 0.4em\relax American Society of Mechanical Engineers, 2017.

\bibitem{chekroun2024mbappe}
R.~Chekroun, T.~Gilles, M.~Toromanoff, S.~Hornauer, and F.~Moutarde, ``{MBAPPE: MCTS-built-around prediction for planning explicitly},'' in \emph{2024 IEEE Intelligent Vehicles Symposium (IV)}.\hskip 1em plus 0.5em minus 0.4em\relax IEEE, 2024, pp. 2062--2069.

\bibitem{wen2024monte}
Q.~Wen, Z.~Gong, L.~Zhou, and Z.~Zhang, ``{Monte Carlo Tree Search for Behavior Planning in Autonomous Driving},'' in \emph{2024 IEEE International Symposium on Safety Security Rescue Robotics (SSRR)}.\hskip 1em plus 0.5em minus 0.4em\relax IEEE, 2024.

\bibitem{karimi2020receding}
S.~Karimi and A.~Vahidi, ``Receding horizon motion planning for automated lane change and merge using monte carlo tree search and level-k game theory,'' in \emph{2020 American Control Conference (ACC)}.\hskip 1em plus 0.5em minus 0.4em\relax IEEE, 2020.

\bibitem{coulom2006efficient}
R.~Coulom, ``Efficient selectivity and backup operators in monte-carlo tree search,'' in \emph{International conference on computers and games}.\hskip 1em plus 0.5em minus 0.4em\relax Springer, 2006, pp. 72--83.

\bibitem{silver2016mastering}
D.~Silver, A.~Huang, C.~J. Maddison, A.~Guez, L.~Sifre, G.~Van Den~Driessche, J.~Schrittwieser, I.~Antonoglou, V.~Panneershelvam, M.~Lanctot \emph{et~al.}, ``{Mastering the game of Go with deep neural networks and tree search},'' \emph{nature}, 2016.

\bibitem{silver2017mastering}
D.~Silver, T.~Hubert, J.~Schrittwieser, I.~Antonoglou, M.~Lai, A.~Guez, M.~Lanctot, L.~Sifre, D.~Kumaran, T.~Graepel \emph{et~al.}, ``Mastering chess and shogi by self-play with a general reinforcement learning algorithm,'' \emph{arXiv preprint arXiv:1712.01815}, 2017.

\bibitem{gonzalez2019human}
D.~S. Gonz{\'a}lez, M.~Garz{\'o}n, J.~S. Dibangoye, and C.~Laugier, ``Human-like decision-making for automated driving in highways,'' in \emph{2019 IEEE intelligent transportation systems conference (ITSC)}.

\bibitem{zhang2020efficient}
L.~Zhang, W.~Ding, J.~Chen, and S.~Shen, ``Efficient uncertainty-aware decision-making for automated driving using guided branching,'' in \emph{2020 IEEE International Conference on Robotics and Automation (ICRA)}.\hskip 1em plus 0.5em minus 0.4em\relax IEEE, 2020, pp. 3291--3297.

\bibitem{nelson}
N.~De~Moura, R.~Chatila, K.~Evans, S.~Chauvier, and E.~Dogan, ``Ethical decision making for autonomous vehicles,'' in \emph{2020 IEEE Intelligent Vehicles Symposium (IV)}, 2020, pp. 2006--2013.

\bibitem{IDM}
A.~Kesting, M.~Treiber, and D.~Helbing, ``Enhanced intelligent driver model to access the impact of driving strategies on traffic capacity,'' \emph{Philosophical Transactions of the Royal Society A: Mathematical, Physical and Engineering Sciences}, vol. 368, no. 1928, pp. 4585--4605, 2010.

\bibitem{moghadam2021autonomous}
M.~Moghadam and G.~H. Elkaim, ``An autonomous driving framework for long-term decision-making and short-term trajectory planning on frenet space,'' in \emph{2021 IEEE 17th International Conference on Automation Science and Engineering (CASE)}.\hskip 1em plus 0.5em minus 0.4em\relax IEEE, 2021.

\bibitem{MOBIL}
A.~Kesting, M.~Treiber, and D.~Helbing, ``General lane-changing model mobil for car-following models,'' \emph{Transportation Research Record}, vol. 1999, no.~1, pp. 86--94, 2007.

\bibitem{rosin2011multi}
C.~D. Rosin, ``Multi-armed bandits with episode context,'' \emph{Annals of Mathematics and Artificial Intelligence}, vol.~61, no.~3, 2011.

\bibitem{exp3}
P.~Auer, N.~Cesa-Bianchi, Y.~Freund, and R.~E. Schapire, ``The nonstochastic multiarmed bandit problem,'' \emph{SIAM journal on computing}, vol.~32, no.~1, pp. 48--77, 2002.

\bibitem{kocsis2006bandit}
L.~Kocsis and C.~Szepesv{\'a}ri, ``Bandit based monte-carlo planning,'' in \emph{European conference on machine learning}.\hskip 1em plus 0.5em minus 0.4em\relax Springer, 2006.

\bibitem{auer2002finite}
P.~Auer, ``{Finite-time Analysis of the Multiarmed Bandit Problem}.''

\bibitem{RTMapsFawzi}
F.~Nashashibi, ``/sup rt/m@ps: a framework for prototyping automotive multi-sensor applications,'' in \emph{Proceedings of the IEEE Intelligent Vehicles Symposium 2000 (Cat. No.00TH8511)}, 2000, pp. 99--103.

\end{thebibliography}

\end{document}